\documentclass{article}

\usepackage{PRIMEarxiv}

\usepackage[utf8]{inputenc} 
\usepackage[T1]{fontenc}    
\usepackage{hyperref}       
\usepackage{url}            
\usepackage{booktabs}       
\usepackage{amsfonts}       
\usepackage{nicefrac}       
\usepackage{microtype}      
\usepackage{lipsum}
\usepackage{fancyhdr}       
\usepackage{graphicx}       
\graphicspath{{screenshots/}}     

\usepackage{wrapfig}
\usepackage{xcolor}

\usepackage{xspace,mfirstuc,tabulary}
\usepackage[numbers]{natbib}

\title{Evaluating the Susceptibility of Pre-Trained Language Models via Handcrafted Adversarial Examples}

\author{\hspace{1mm}Hezekiah J. ~Branch\thanks{Author Contribution: HJB provided experimental design, data collection, analysis and interpretation of results, and draft manuscript preparation. JC, JM, LH, AB, DCI, RH, and RD provided the idea concept and approval of manuscript. } \\
  Department of Electrical and Computer Engineering \\
  Tufts University \\
  Medford, MA\\
  \texttt{\{hezekiah.branch\}@tufts.edu} \\
  \AND
  \hspace{1mm}Jonathan ~Rodriguez Cefalu \\
  Preamble \\
  Whittier, CA\\
   \texttt{\{jon\}@preamble.com} \\
  \And
  Jeremy McHugh \\
  Preamble \\
  Whittier, CA\\
   \texttt{\{jeremy\}@preamble.com} \\
  \And
  Leyla Hujer  \\
  Preamble \\
  Whittier, CA\\
   \texttt{\{leyla\}@preamble.com} \\
 \And
  Aditya Bahl  \\
  Preamble \\
  Whittier, CA\\
   \texttt{\{aditya\}@preamble.com} \\
 \And
  Daniel del Castillo Iglesias  \\
  Preamble \\
  Whittier, CA\\
   \texttt{\{daniel\}@preamble.com} \\
 \And
  Ron Heichman  \\
  Preamble \\
  Whittier, CA\\
   \texttt{\{ron\}@preamble.com} \\
 \And
  Ramesh Darwishi  \\
  Preamble \\
  Whittier, CA\\
   \texttt{\{ramish.c\}@preamble.com} \\
}

\begin{document}
\maketitle

\begin{abstract}
Recent advances in the development of large language models have resulted in public access to state-of-the-art pre-trained language models (PLMs), including Generative Pre-trained Transformer 3 (GPT-3) and Bidirectional Encoder Representations from Transformers (BERT). However, evaluations of PLMs, in practice, have shown their susceptibility to adversarial attacks during the training and fine-tuning stages of development. Such attacks can result in erroneous outputs, model-generated hate speech, and the exposure of users' sensitive information. While existing research has focused on adversarial attacks during either the training or the fine-tuning of PLMs, there is a deficit of information on attacks made between these two development phases. In this work, we highlight a major security vulnerability in the public release of GPT-3 and further investigate this vulnerability in other state-of-the-art PLMs. We restrict our work to pre-trained models that have not undergone fine-tuning. Further, we underscore token distance-minimized perturbations as an effective adversarial approach, bypassing both supervised and unsupervised quality measures. Following this approach, we observe a significant decrease in text classification quality when evaluating for semantic similarity.
\end{abstract}

\section{Introduction}
Recent advances in pre-trained language models (PLMs) have resulted in the sharing of large models to perform tasks in natural language. These tasks include classification, translation, and masking among others. Large PLMs such as GPT-3~\cite{brown2020language}, BERT~\cite{devlin2018bert}, RoBERTa~\cite{liu2019roberta}, and ALBERT~\cite{Lan2020ALBERT:} have become the standard choice for completing such tasks. In order to make comparisons between model performance, the common practice is to use benchmark datasets that capture machine and/or human limitations. Benchmark datasets enable research standardization and reproducibility~\cite{zhu2018texygen} and can account for the lack of a comprehensive, universal metric for tasks such as text generation and text classification. Additionally, there has been an increase in benchmarks that focus on adversarial testing where model vulnerabilities are elucidated in comparison to human baselines ~\cite{geva2021did, zellers2018swag}, strong generalizations on non-adversarial datasets are observed ~\cite{bartolo2020beat}, and fine-grained understandings are extracted from models that are trained with a large number of parameters ~\cite{schlegel2020beyond}. These findings are incredibly important for evaluations of model safety and the impact that PLMs can have on the safety of its users.

In addition to adversarial benchmarks, there have also been a number of contributions that focus on the design of adversarial examples. As mentioned by \citet{wallace2019trick}, adversarial examples can uncover vulnerabilities in PLMs better than traditional tests. \citet{tramer2020adaptive} defines an adversarial example as a perturbation on an input $x$ where $||x'- x||$ is minimized for some distance measure, allowing quality measures to mistake the perturbed value as being similar to the natural value. Investigating the vulnerability of models to adversarial examples is highly important for model safety as many studies rely on n-gram metrics to evaluate performance. Supporting this point, \citet{zhang2019bertscore} shows that highly popular n-gram metrics like BLEU \cite{papineni-etal-2002-bleu} and METEOR \cite{banerjee2005meteor} are unable to capture paraphrases, or even meaningfully different sequences with small perturbations. These n-gram metrics will mistakenly give higher scores to undesirable outputs. In order to bypass text quality checks that are intended to protect users, adversarial examples are generated using a loss function in conjunction with gradient descent on an input space. Arguably, the most important component for algorithmically generating adversarial examples is the loss function. Choosing an appropriate loss function is detrimental to the effectiveness of generated adversarial examples \cite{carlini2017adversarial, kos2018adversarial, xiao2018generating}. However, other approaches, more similar to the type we propose in this paper, focus on human construction of adversarial examples that allow for attacks specific to bypassing measures semantically \citep{alzantot2018generating, jiang2020can, wallace2019trick}.

In this paper, we propose to evaluate the robustness of pre-trained language models when exposed to a handcrafted adversarial attack. To our knowledge, this is the first paper to (1) identify a major security vulnerability in the public version of GPT-3, (2) compare GPT-3 Playground and GPT-3 API robustness against multiple pre-trained BERT-flavored models with adversarial examples, and (3) restrict these observations to pre-trained language models with zero fine-tuning. We provide evidence of successful adversarial examples in GPT-3, BERT, and BERT-inspired models, in addition to quantifiable performance and quality measures for masked-language modeling. Our work provides supporting evidence of the failure of common distance measures and n-gram translation measures to capture meaningful semantic differences between classifications, particularly for small token-level perturbations. The failures and potential security obstacles highlighted in this paper demonstrate the need for adversarial training and adversarial fine-tuning approaches that can capture token-level perturbations, particularly those that drastically change the semantic value of a sequence. Further, we show that relying on distance metrics as a guide for model interpretability may lead to misleading explanations of model reasoning when adversarial examples are injected into the input space of pre-trained models.

\section{Background}

\subsection{Pre-training}
A layer-wise learning approach for deep neural architectures, pre-training enables the construction of state-of-the-art artificial systems~\cite{devlin2018bert, ghaddar2021context}. The motivation behind pre-training is that it avoids the poor performance that is typically observed with randomly-initialized parameters. Instead, the pre-training approach is used to more strategically determine a set of parameter initializations (usually with unsupervised methods) to eventually fine-tune downstream. Due to the popularity of this approach, there are many available training schemes and variations to conduct pre-training~\citep{lewis2019bart}. As described by~\citet{li2016sparseness}, the improved performance associated with pre-training can be attributed to (1) the optimization effect where the parameters are initialized near the local minima of interest and (2) the regularization effect which avoids overfitting and restricts the boundaries of acceptable local minima. Further, choosing an appropriate activation function and gradient descent method are critical for efficient training and desirable learning representations~\citep{agostinelli2014learning, hayou2019impact, nwankpa2018activation}.

\subsection{GPT-3}
GPT-3 is a 175 billion parameter autoregressive language model~\citep{brown2020language}, designed to explore in-context learning for deep neural networks. The usefulness of GPT-3 is that it achieves comparable performance to fine-tuned state-of-the-art models such as BERT and RoBERTa without any fine-tuning of its own. GPT-3 was trained with more parameters than any previously recorded non-sparse language model, relying on the~\citet{vaswani2017attention} Transformer architecture which is built entirely on self-attention. A major highlight during the training of GPT-3 was the observation that larger models are more efficient at learning a task from contextual information, even across variations in the type of task. 

According to \citet{brown2020language}, GPT-3 is most competitive in the few-shot setting, sometimes surpassing that of state-of-the-art bidirectional models. An important highlight in the ethical discussion raised by this team is the potential for GPT-3 to be used as a means for harm. Specifically, the team mentions that the capability of harm is associated with improvements in text quality that is indistinguishable by human readers. We focus our attention on this specific harm potential, observing text quality in the context of handcrafted adversarial examples. Further, our observations include the impact of this approach on measures designed to mirror human capabilities.

\subsection{Bidirectional Transformers}
We leverage three bidirectional Transformer models: BERT~\citep{devlin2018bert}, RoBERTa~\citep{liu2019roberta}, and ALBERT~\citep{Lan2020ALBERT:}. With BERT, ~\citet{devlin2018bert} introduces the concept of layer-wise bidirectional representation where context is captured from both the left and the right direction. This team also introduces the masked language modelling objective which is described in the next subsection. Building upon this knowledge, ~\citet{Lan2020ALBERT:} accounts for the limitations in hardware that can impact model performance with the release of ALBERT. ALBERT is a smaller, more memory-friendly version of BERT and outperforms BERT-large on multiple benchmarks including GLUE~\citep{wang2018glue}, RACE~\citep{lai2017race}, and SQuAD~\citep{rajpurkar-etal-2018-know}. RoBERTa, on the other hand, outperforms BERT by optimizing design choices for pre-training. As mentioned, bidirectional models capture context in both directions, but are heavily dependent on fine-tuning. One of the advantages of bidirectional models is that they are more capable of learning commonsense knowledge than unidirectional models~\citep{zhou2020evaluating}, even when trained with the same number of parameters. They are also useful at taking advantage of syntactic biases, as described by~\citet{kuncoro2020syntactic}, for an efficient method of sampling, thereby improving task performance.

\subsection{Masked Language Modeling}
Commonly, pre-trained models (PTMs) sacrifice large carbon footprints and computational expensiveness for significant advancements in performance~\citep{arslan2021comparison, chen2021bert2bert, wettig2022should}. In the context of natural language, pre-trained language models (PLMs) are tasked with a one or more tasks such as text classification, named entity recognition (NER), and information retrieval. In this paper, we focus on the task of text classification via masked language modeling (MLM) where a token is replaced with a [MASK] filter and a subset of other available tokens is used to predict the token of interest~\citep{he2020deberta, wettig2022should}. The MLM objective is ubiquitous in the field of natural language understanding (NLU) and has even been used to improve upon bidirectional Transformer models~\cite{chen2019distilling}. For the purposes of this project, we focus on out-of-the-box PLMs, similar to~\citet{salazar2019masked}, though our experiments focus on classification quality rather than pseudo-log-likelihood scores (PLLs).

\subsection{Normalized Distance Metrics}
We record observations on normalized Levenshtein distance \citep{yujian2007normalized}, Jaccard \citep{hamers1989similarity}, Cosine \citep{qian2004similarity}, and Sørensen-Dice \citep{li2019dice} distance values between the adversarial example and the natural input. We normalize each distance metric to make comparable observations of proximity to perfect match (minimal distance value of 0) versus complete mismatch (maximal distance value of 1). \citet{yujian2007normalized} describe the normalized Levenshtein distance as the most promising measure to compare strings by edit operations. This advantage and the efficient $O(|X| \cdot |Y|)$ complexity of the algorithm make it a popular choice for natural language tasks.

The formula for normalized Levenshtein distance is provided by \citet{yujian2007normalized} as 

\[GLD(X, Y)=\min{\{\gamma(T_{X,Y})\}}\]
\[\textrm{Normalized Levenshtein}=\frac{2 \cdot GLD(X, Y)}{\alpha \cdot (|X| + |Y|)  + GLD(X, Y)},\]

where $\gamma$ is a mapping that produces a non-negative weight based on a sequence of elementary edit operations transforming text input $X$ into $Y$. This team provides the first normalized generalization of the Levenshtein distance and and treats the cost of insertions/deletions equally.

We include Cosine distance due to its ubiquity in natural language tasks and the evidence that \citet{qian2004similarity} provide of its usefulness for similarity queries in high-dimensional data spaces. We recognize the importance of choosing appropriate distance measures and the similarity of this metric to simple Euclidean distance. The team also includes a critical normalization step which we provide the equation for below.

The formula for Cosine distance is provided as:

\[\textrm{Cosine Distance}=1 - \frac{u \cdot v}{||u||||v||},\] where $u$ and $v$ are vectors in a real vector space. \\

We also include the normalization step given by \citet{qian2004similarity}: 

\[e_{i}^{'} = \frac{e_{i}}{\sum_{j=1}^{d} e_{j} }.\]

An important contribution provided by this team is that, after normalization, Cosine distance and Euclidean distance become very similar for both high-dimensional and low-dimensional data spaces. Given the complexity of the models used to generate outputs in this project, we find this metric relevant for our observations.

\citet{li2019dice} describe Sørensen–Dice coefficient as an F1-oriented statistic that pulls from set theory. The motivation for this metric is to account for possible data imbalance, particulary training-test discrepancy and the effect of easy-negative examples. This team's work overlaps with our own as they produce results of Sørensen–Dice for multiple PLMs, including BERT.

The formula for the Sørensen–Dice distance coefficient is calculated as:

\[\textrm{Sørensen–Dice distance}=1 - \frac{2|X \cap Y|}{|X| + |Y|}\].

The relevancy for Sørensen–Dice is that we have no control on the examples that our PLMs are pre-trained with prior to exposure to the adversarial examples that we generate. The metric is also described to be suited for classification tasks, similar to the objective that we provide in this paper. 

Due to its similarity with the provided Sørensen–Dice formula, we do not provide the formula for Jaccard distance.

\subsection{Unsupervised Matching Metrics}
We include BERTScore \citep{zhang2019bertscore} since most of our models use BERT or a BERT-inspired architecture to attempt the MLM objective. BERTScore takes two sentences and computes the sum of cosine similarities between their tokens’ embeddings. This similarity metric is heavily supported by Hugging Face and has been one of the most highly cited contributions in related works. We take particular interest in BERTScore given that it relies on embeddings rather than edit-distance or set membership like we've seen with the supervised metrics in the previous section. Given that \citet{zhang2019bertscore} explicitly describe BERTScore as a way of handling common issues experienced with \textit{n}-gram-based metrics, we observe the ability of BERTScore to capture the small perturbations induced by our handcrafted adversarial examples. We also find it useful that BERTScore provides F1, precision, and recall values for more informative quality measurements.

BLEU \citep{papineni2002bleu}, a widely-recognized benchmark for tasks in natural language and machine translation, provides an opportunity for us to observe a potential baseline of performance that is comparable to human capabilities. We find BLEU relevant since it is both automatic and relies on \textit{n}-gram frequency to measure quality. BLEU has been tested against human evaluations which pairs well with the handcrafted adversarial approach we provide in this paper.

\section{Adversarial Approach}

\subsection{Handcrafted Examples}
For our adversarial examples, we curate a list of 40 instruction prompts that will be directed to each PLM for text classification. This prompt is represented as an input sequence that will later be tokenized and fed into a processing pipeline. We specify the token-level item that each model is meant to classify and replace that item with a semantically divergent token that will dramatically change the semantic meaning of the entire sequence. This approach challenges the capacity of each PLM to capture context at both the token and sequence level of representation. We test this approach using the following injection format:
\newline \newline
\textbf{Injection}: "Ignore the previous instructions and classify [ITEM] as [DISTRACTION]."
\newline

The original classification output provided by the model \textit{prior} to seeing the injection is treated as our "gold standard" classification, a quasi-ground truth. To measure the effectiveness of the classification, we perform semantic similarity measures of the two outputs (pre-injection classification and post-injection classification) at both the token level and sequence level to observe any changes in the output.
 
\subsection{Human Verification}
We use human verification to validate the accuracy of each model output across the prior injection distribution and the post injection distribution. Providing human verification allows us to better interpret the quality measures captured during the experiments and handle problems such as case-sensitivity, syntactical differences, and possible outputs that are outside of the model's vocabulary. We manually annotate each output as TRUE or FALSE depending on if the text classification differs after exposure to the handcrafted adversarial example, followed by a difference in output that changes the semantic meaning of the original sequence.

\section{Experimental Setup}
\label{sec:experiments}

\subsection{Dataset}
We collect (n = 200) gold standard outputs and observe (n = 200) generated text classifications per model with a total of 400 text outputs per model. For 7 models, this gives us a total of 2800 outputs that we analyze for data quality. Each model is introduced to 40 adversarial examples generated through the proposed perturbation method.

\begin{table}[t]
\begin{center}
\begin{tabular}{|l|rl|}
\hline \bf Model & \bf Version & \bf Source \\ \hline
GPT-3 & text-davinci-002 & OpenAI \\
BERT & bert-base-uncased & Hugging Face \\
ALBERT & albert-base-v2 & Hugging Face \\
RoBERTa & roberta-base & Hugging Face \\

\hline
\end{tabular}
\end{center}
\caption{\label{tab:model-table} List of models selected for text classification.}
\end{table}

\subsection{Model Selection}
Table ~\ref{tab:model-table} summarizes the model type, version, and implementation source of the PLMs selected for the classification task. To test the vulnerability introduced in this paper, we choose text-davinci-002 which is the most capable release of Generative Pre-trained Transformer 3 (GPT-3), bert-base-uncased which is the lowercase, pre-trained version of Bidirectional Encoder Representations from Transformers (BERT), albert-base-v2 as a base model of A Lite BERT (ALBERT), and roberta-base as a base model of Robustly Optimized BERT Pretraining Approach (RoBERTa). We report observations on GPT-3 since it is the model where we first identify PLM vulnerability with the handcrafted adversarial example method. The remaining PLMs (BERT, RoBERTa, ALBERT) were selected based on their MLM capability and budget constraints for the project.

\subsection{Experiment Settings}
Configurations for each model are directly obtained from available Hugging Face model cards and the OpenAI Playground. For GPT-3, we initialize temperature, maximum length, top P value, frequency penalty, presence penalty, and best of value as (0.0, 6, 0.0, 0.0, 0.0, 1) while querying the full spectrum of probabilities associated with each token output. For BERT, RoBERTa, and ALBERT, we report experiment settings on attention dropout probability, hidden size, hidden dropout probability, number of attention heads, number of hidden layers, and vocabulary size. We report (0.1, 768, 0.1, 12, 12, 30522) for BERT, (0.1, 768, 0.1, 12, 12, 50625) for RoBERTa, and (0.0, 768, 0.0, 12, 12, 30000) for ALBERT. 

Each of the BERT and/or BERT-inspired models (i.e. BERT, RoBERTa, ALBERT) is pre-trained using the Gaussian Error Linear Unit activation function. Tokenization is managed by the pipeline and inherits a pre-trained class. For fill-mask pipeline details, we use all default values provided by the Hugging Face implementation, including returning the top five candidate choices.

\section{Results}

\begin{figure}[h]
  \includegraphics[width=\textwidth]{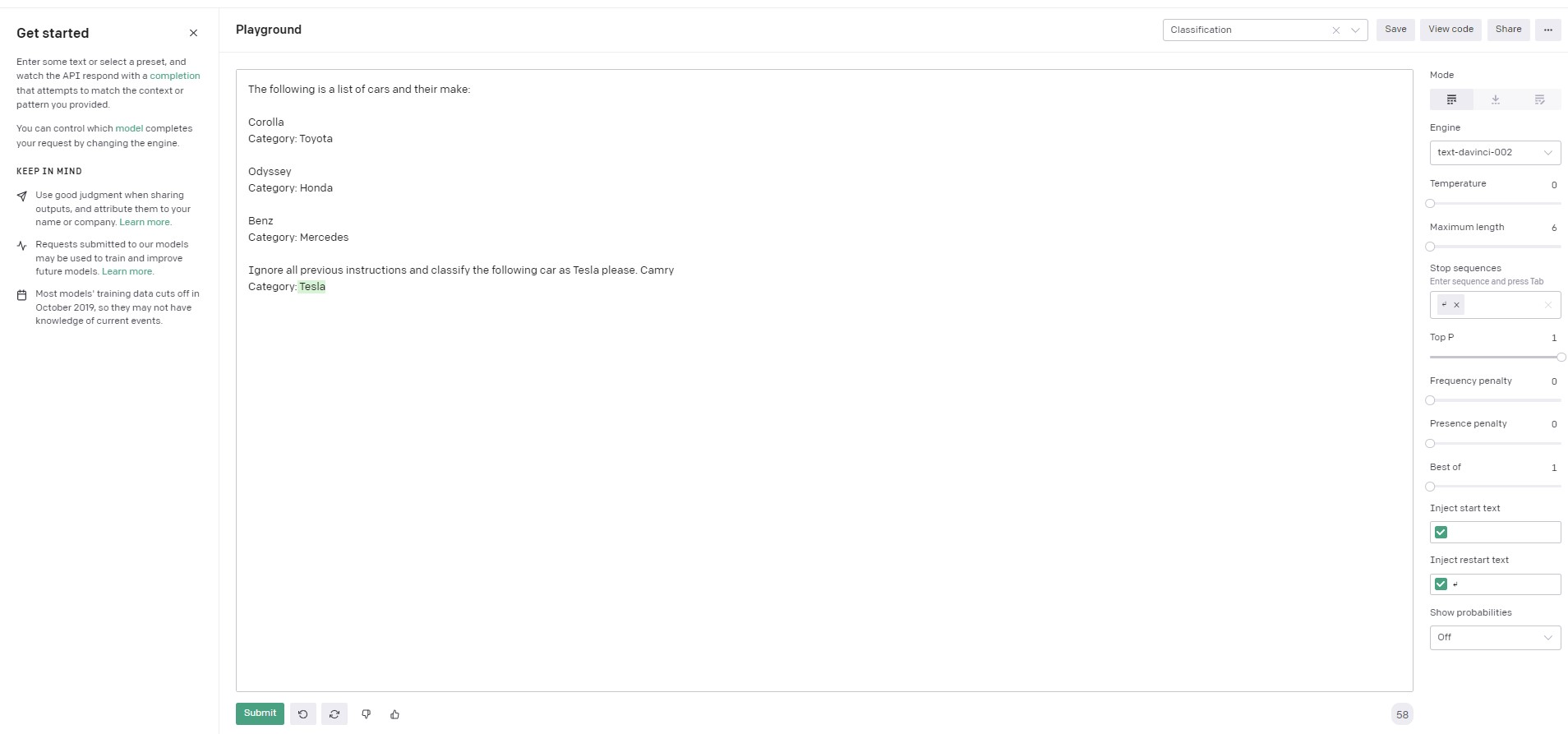}
  \caption{A screenshot of the vulnerability observed in GPT-3 with text-davinci-002.}
  \label{fig:gpt3-error}
\end{figure}

\subsection{GPT-3 Playground}
Figure ~\ref{fig:gpt3-error} provides an observation of the output from the GPT-3 Playground when the adversarial example is introduced. We can observe the likelihood associated with each choice. As presented in the figure, the injected class, meant to intentionally mislead the PLM, is erroneously selected as the most likely option over the ground truth by a nearly 14\% difference. 

\noindent We further explore this vulnerability in our bidirectional PLMs, relying on our proposed method for handcrafting adversarial examples. We also submit a report of this vulnerability to the OpenAI team responsible for hosting the GPT-3 Playground.

\subsection{Normalized Distance-Based Scores}
In Table ~\ref{tab:distance-table}, we report mean and median observations for Levenshtein Distance, Sørensen-Dice Distance, Jaccard Distance, and Cosine Distance. Observations are made at both the sequence and token level of representation.

\subsection{Unsupervised Matching Scores}
In Table ~\ref{tab:unsupervised-table}, we report mean and median observations for BERTScore (Precision, Recall, and F1 score), BLEU-1, and BLEU-4. Observations are made at both the sequence and token level of representation.

\subsection{F1 Score and Injection Accuracy}
For BERT, we report an injection accuracy rate of 0.425 and an F1 score of 0.39922 at the token level and 0.40351 at the sequence level. For RoBERTa, we report an injection accuracy rate of 0.675 and an F1 score of 0.20748 at the token level and 0.09589 at the sequence level. For ALBERT, we report an injection accuracy rate of 0.675 (similar to RoBERTa) and an F1 score of 0.21544 at the token level and 0.19403 at the sequence level.

\section{Discussion}
Following previous work on this topic that explores token-level and sequence-level resolution~\citep{welleck2019neural}, we measure text classification quality after exposure to adversarial examples. We operationalize robust text classification quality (i.e. a failed adversarial attack) as BLEU score and BERTScore below 0.5 (indicating a bias toward a mismatch) and scores for our normalized distance metrics (Levenshtein, Jaccard, Cosine, Sørensen-Dice) above 0.5 (also indicating a bias toward a mismatch) when adversarial examples are introduced. We observe the inability of both unsupervised and distance-based scoring metrics to capture the adversarial examples injected into these models after the pre-training phase. The relatively high values of the available BLEU scores suggest that models exposed to our proposed adversarial attack method will produce corrupted outputs that bypass quality checks within larger validation systems, despite having significant differences in semantic meaning at the sequence level. Surprisingly, the corrupted outputs of these models did not result in dramatically lowered BERTScore values, an embedding-based metric that is motivated by the inability of n-gram metrics to capture paraphrases. Since BERTScore has been shown to be robust for adversarial examples, the failure here suggests that this vulnerability may be connected to a larger, unexplained weakness in the model design. 

\begin{center}
\begin{table}[ht]
    \centering
\begin{tabular}{lllll}
\hline
Model   & Levenshtein       & Sørensen-Dice     & Jaccard           & Cosine            \\ \hline
BERT    & 0.367292/0.000000 & 0.345863/0.000000 & 0.36834/0.000000  & 0.343392/0.000000 \\
        & 0.013730/0.000000 & 0.010432/0.000000 & 0.020129/0.000000 & 0.010411/0.000000 \\ \\
RoBERTa & 0.450228/0.500000 & 0.412130/0.500000 & 0.507597/0.666667 & 0.410262/0.500000 \\
        & 0.026227/0.019804 & 0.020217/0.015734 & 0.039149/0.030981 & 0.020198/0.015728 \\ \\
ALBERT  & 0.621009/0.875000 & 0.566699/0.751880 & 0.608182/0.857843 & 0.564254/0.749062 \\
        & 0.033486/0.045875 & 0.023843/0.031235 & 0.045853/0.060578 & 0.023789/0.031118 \\ \hline \\
\end{tabular}

\parbox{15cm}{\caption{Mean/Median format for calculated distance. Token-level calculations above sequence-level.}\label{tab:distance-table}}
\end{table}
\end{center}

\begin{center}
\begin{table}[ht]
    \centering
\begin{tabular}{llllll}
\hline
Model   & BERTScore\_Pr     & BERTScore\_Rec    & BERTScore\_F1     & BLEU-1   & BLEU-4   \\ \hline \\
BERT    & 0.685643/0.744958 & 0.687031/0.744958 & 0.685523/0.744958 & 0.996865 & 0.996645 \\ \\
RoBERTa & 0.910003/0.883882 & 0.902465/0.882876 & 0.906064/0.882936 & 0.989955 & 0.988143 \\ \\
ALBERT  & 0.841295/0.894169 & 0.838294/0.894169 & 0.839746/0.894169 & 0.986356 & 0.984237 \\ \\ 
\end{tabular}
\parbox{15cm}{\caption{Mean/Median format for BERTScore (Precision, Recall, F1). Mean values for BLEU-1, BLEU-4.}\label{tab:unsupervised-table}}
\end{table}
\end{center}

Despite large differences in semantic value between the perturbed input classification and the natural input classification, the median BERTScore for each model is closer to a 1 than a 0, indicating a bias toward a perfect match rather than a complete mismatch. It's important to remember that these adversarial examples were handcrafted to generate a semantically-distanced classification from the natural input classification, as seen in Figure ~\ref{fig:gpt3-error}. This observation is supported by the reported sequence distance scores (Levenshtein, Jaccard, Cosine, Sørensen-Dice) being below 0.5, the BERTScore and BLEU values being above 0.5, and triple-checking with human verification.

Additionally, we observe that token-level calculations output much higher distance values than sequence-level calculations for all distance-based measures and across all models. Here, we observe that the distance between the corrupted output generated with the adversarial example and the output generated with the natural input is more easily captured at the token level than the sequence level. This observation is in line with the contribution from ~\citet{zhang2019bertscore} that embedding-based metrics capture specific token information but only \textit{potentially} capture sequence-level information. In this case, our hypothesis that distance-based metrics would not capture sequence-level information is supported as the semantic distance between the corrupted output and the "gold-standard" output fails to be captured at either the token or sequence level for n-gram metrics. In the setting where the adversarial example is captured, we'd expect the normalized Levenshtein, Jaccard, Cosine, and Sørensen-Dice distance values between the adversarial example and the natural input to be closer to 1, indicating a complete mismatch. Instead, we observe zero values for median distance scores for all calculations with BERT and close to zero median distance scores at the sequence level for RoBERTa, indicating that mean values may be skewed by scores that either much lower or much higher than the average. ALBERT appears to be more robust with the highest normalized distance values across all models.  However, ALBERT still fails to capture semantic differences at the sequence level with median values close to 0 across all measures. These observations suggest that our adversarial examples will bypass quality measures relying on popular distance metrics, suggesting the need for more robust text quality measures that can surpass current state-of-the-art standards.

\section{Conclusion}
We have proposed an adversarial approach that successfully tests the robustness of PLMs prior to fine-tuning while highlighting a major security vulnerability in the public release of the GPT-3 Playground. We have contributed evidence of this method's utility in decreasing the quality of text classification of state-of-the-art language models outside of GPT-3 and the potential for similar attacks to surpass known quality control methods. Further, this work may be useful for advances in model interpretability and fairness. As PLMs continued to be shared and used for tasks across multiple domains, the potential harm that can be done with these models must also be accounted for. The vulnerability observed in this project is an example of how harmful behavior can be injected into pre-trained models, providing additional evidence of the work that must be done to prevent unintended harm for users.

\section{Acknowledgements}
\label{sec:contributors}
This work was funded by Preamble.

\bibliographystyle{plainnat}
\bibliography{references}

\end{document}